\documentclass{article}
\usepackage[preprint]{neurips_2026}

\usepackage[utf8]{inputenc} 
\usepackage[T1]{fontenc}    
\usepackage{hyperref}       
\usepackage{url}            
\usepackage{booktabs}       
\usepackage{amsfonts}       
\usepackage{nicefrac}       
\usepackage{microtype}      
\usepackage{xcolor}         
\newcommand{\changedcolor}{\color{black}}
\DeclareRobustCommand{\changed}[1]{{\changedcolor#1}}

\usepackage{graphicx} 
\usepackage{epstopdf}

\usepackage{amsmath}
\usepackage{amssymb}
\usepackage{amsthm}
\usepackage[capitalize,noabbrev]{cleveref}
\usepackage{mathtools}
\usepackage{qtree}
\usepackage{soul}
\usepackage{subcaption}
\DeclareCaptionFont{changed}{\changedcolor}
\usepackage[textsize=tiny]{todonotes}

\usepackage{amsmath,amsfonts,bm}

\def\eqref#1{equation~\ref{#1}}

\def\1{\bm{1}}

\DeclareMathAlphabet{\mathsfit}{\encodingdefault}{\sfdefault}{m}{sl}
\SetMathAlphabet{\mathsfit}{bold}{\encodingdefault}{\sfdefault}{bx}{n}

\definecolor{mypink}{RGB}{255, 105, 180}

\title{Towards Data\changed{\mbox{-}}Efficient\changed{,} Real\changed{\mbox{-}}Time World Models}

\author{%
  \textbf{Raluca Stevenson} \quad
  \textbf{David Bignell} \quad
  \textbf{Jennifer C.~White} \quad
  \textbf{Tabish Rashid}\thanks{Now at Google DeepMind.}
  \\
  \textbf{Emanuele Aiello}\thanks{Now at Prometheus.} \quad
  \textbf{Mariya Hendriksen}\thanks{Now at the University of Oxford.} \quad
  \textbf{Junha Hyung}\thanks{Now at the Korea Advanced Institute of Science and Technology (KAIST).}
  \\
  \textbf{Sadegh Aliakbarian}\thanks{Now at Google.} \quad
  \textbf{Abdelhak Lemkhenter}\thanks{Now at Isomorphic Labs.} \quad
  \textbf{Marko Tot}\thanks{Now at Iconic.}
  \\
  \textbf{Katja Hofmann} \quad
  \textbf{Sarah Parisot}
  \\[0.5em]
  Microsoft, Cambridge, United Kingdom
}

\begin{document}

\maketitle

\begin{abstract}
State-of-the-art world models have been used to produce sequences of gameplay that accord with provided user-input actions.
However, high quality, consistent gameplay generation often comes at the cost of inference speed, making real-time interactive play and ideation challenging. 
The substantial quantities of data used to train state-of-the-art models have also made custom training of these models appear impractical, limiting many of their potential uses in creative fields.
In this work we demonstrate a major step towards efficient world models: achieving both fast high-quality inference, and improved data efficiency.
This is done by replacing the next-token prediction paradigm with a two-stage discrete diffusion-based approach, introducing a lightweight refinement transformer which carries out iterative masked predictions. Building on an autoregressive world model capable of generating highly consistent and complex sequences over minutes \citep{Kanervisto2025}, we achieve substantial model speed-up (79x) \changed{with a modest quality trade-off}.
Additionally, we demonstrate that this model can be trained to reproduce high-quality, consistent gameplay using a dataset containing approximately one day (28 hours) of gameplay data.
Our results demonstrate progress towards removing barriers to the use of world models in settings where real-time inference and data efficiency are essential.
\end{abstract}

\section{Introduction}
In recent years, impressive progress has been achieved towards high quality video generation. Large scale models \citep{sora, veo} trained on real-world footage, video games, and interactive 3D environments have showcased a potential to learn rich and complex environments and interactions. While text remains the main mode of interaction with video generation models, there is growing interest in action-controlled models (e.g. using navigation keys, game controller actions), often referred to as interactive video models or \emph{world models} \citep{genie3, oasis, matrix}. Potential applications for such world models include self-driving simulation \citep{gaia}, robotics \citep{cosmos}, and supporting creative ideation and game development \citep{Kanervisto2025}. 
In the context of video gaming, models trained on large scale single game environments have demonstrated an ability to learn and faithfully reproduce game maps and physics, complex interactions and character abilities in a responsive environment \citep{gamengenx, diamond}. 

Despite these advances, current models suffer from drawbacks that limit their practical use: many world models
struggle to generate long, high-fidelity, consistent gameplay in real-time. Models capable of real-time generation are often limited to simple interactions (e.g. navigation) and video quality degrades after a minute of gameplay \citep{diamond}. In contrast, autoregressive models like WHAM \citep{Kanervisto2025} are capable of generating long videos with high fidelity visuals and complex interactions, but 
take minutes to generate a few video frames. This is due to the inherently sequential nature of next-token prediction in a high-dimensional discrete space: generating a single frame requires hundreds of sequential forward passes through a large transformer. 
A model capable of real-time, high-fidelity, consistent gameplay generation would greatly expand the potential of world models in practical settings, providing instant responses to user inputs.

Additionally, most state-of-the-art world models are trained on very large quantities of data, making data collection costs and compute requirements prohibitive.
As a result, potential users of world models are limited to publicly available models and APIs, with world model training on custom environments generally seen as unattainable for an individual user or small creative team.
Some of the proposed applications for world models have focused on their uses for creatives, but this avenue is currently blocked, in part because of the large data requirements of a custom model.
In this work, we aim to take steps towards democratisation of world model technology by addressing both of these issues, with the aim of pushing towards world models that are more efficient, both in terms of inference speed and data requirements.

To address the limitations of next-token prediction, we turn to the underexplored paradigm of discrete diffusion \citep{maskgit} for world models. However, discrete diffusion alone is not enough to achieve real-time interactivity (see Figure \ref{fig:fps}) as iterative feed-forward passes remain costly. Instead, we propose to decompose world modelling in two components: 1) a large backbone transformer modelling temporal dependencies which predicts the first image draft in a single forward pass, and 2) a lightweight discrete diffusion transformer
which iteratively refines this prediction. This formulation, combined with a register-based context compression that compresses long context histories into compact representations,  preserves the expressivity of a large transformer with much better scaling to diffusion steps. 
Our experiments show that this design, 
instantiated by WHAM-RT (WHAM Real Time),  
achieves more than $79\times$ speed-up over WHAM with limited quality loss,
maintaining its ability to generate coherent, long horizon videos with complex interactions at 17 frames per second.

To address data efficiency, we demonstrate that \changed{the WHAM-RT dynamics model can be trained from scratch} on a carefully curated dataset comprising one day of gameplay in one game environment. We demonstrate that we can train a high-quality model with comparable and even more stable performance to models trained on the full dataset, effectively reducing data requirements from 1 year to approximately 1 day ($28$ hours) of gameplay. While this subset focuses on a narrower distribution (one character instead of $13$), our results show that it is possible to train high quality models efficiently, taking steps towards enabling creatives to train models on custom environments with limited compute budgets.

To summarise, this work makes two key contributions with the aim of moving world models towards real-time generation and data-efficient training. 
Our contributions are the following:
\begin{enumerate}

    \item We introduce an efficient discrete world model formulation to enable real-time play and interaction for users. Our two-stage architecture leverages discrete diffusion as a lightweight refinement task, in conjunction with a one shot large backbone transformer modelling temporal dependencies. Our resulting WHAM-RT achieves up to $17$ frames per second \changed{on a single H100} and can generate consistent and accurate gameplay for at least 5 minutes.

    \item We demonstrate we are able to train \changed{the WHAM-RT dynamics model from scratch} on approximately 1 day ($28$ hours) of gameplay only, via intentional curation of a high quality data subset, dramatically reducing data requirements from approximately 1 year to 1 day of gameplay. Our experiments show that, on this constrained distribution, models trained on the small dataset achieve better stability over long horizons.

\end{enumerate}

With this work, we aim to highlight the possibility of small and intentional dataset curation for creative prototyping, allowing creators and researchers to train custom models with more modest data and compute budgets.
We hope to motivate future work to make further efforts to achieve high-quality gameplay generation by leveraging a limited amount of data.
\section{Related Work}


\noindent\textbf{Video Generation.}
Video generation has evolved along two dominant paradigms, each with distinct architectures and trade-offs. Diffusion-based approaches, particularly Latent Diffusion Models \citep{rombach2022high} with Diffusion Transformers (DiT) \citep{peebles2023scalable}, achieve strong perceptual quality and temporal stability \citep{ho2022video, blattmann2023stable, esser2024scaling, wan2025wan}. However, their iterative denoising process requires multiple function evaluations, fundamentally limiting inference speed. \changed{Approaches to accelerating diffusion-based approaches include distillation and consistency models \citep{salimans2022progressive,song2023consistency,matrix}, adversarial post-training \citep{sauer2024adversarial,sauer2024fast,lin2025autoregressive}, and efficient model blocks and sparse attention \citep{sanavideo,zhang2025vsa}.}

Autoregressive approaches, inspired by large language models \citep{minaee2024large}, discretize video into token sequences using VQ-VAE or VQ-GAN \citep{van2017neural, esser2021taming} and generate them with causal transformers. Models like VideoGPT \citep{yan2021videogpt}, VideoPoet \citep{kondratyuk2024videopoet} and WHAM \citep{Kanervisto2025} demonstrate precise sequence control but face two critical limitations: sequential decoding bottlenecks that prevent real-time inference, and error accumulation over long horizons. Recent works on non-autoregressive masked prediction, such as \changed{MaskGIT}, MAGVIT and VAR \citep{maskgit, yu2023magvit, VAR}, explore parallel token generation to address these bottlenecks. We take inspiration from these approaches to increase the generation efficiency of autoregressive discrete world models \citep{Kanervisto2025}.
\changed{Lastly, MAR and HART combine autoregressive modelling with lightweight continuous diffusion heads to improve image generation quality and efficiency \citep{MAR,HART}. While architecturally related, WHAM-RT uses a lightweight discrete diffusion transformer for real-time world modelling.}


\noindent\textbf{World Modelling.}
Generative world models go beyond \changed{video generation} by requiring accurate, per frame responsiveness to control signals while maintaining physical plausibility and long-range consistency. Domain-specific progress includes autonomous driving simulators (GAIA-2) \citep{gaia}, interactive gameplay generators (the Genie family) \citep{genie}, and controllable game environments (WHAM) \citep{Kanervisto2025}. Other efforts like WorldMem \citep{xiao2025worldmem} add explicit memory to diffusion backbones to sustain coherence and GameNGen couples reinforcement learning with diffusion for gameplay generation \citep{valevski2025diffusion}.

\changed{Real-time interactive generation has been demonstrated using pixel-space diffusion in DIAMOND \citep{diamond} and continuous-latent diffusion in Oasis \citep{oasis} and Matrix-Game 2.0 \citep{matrix}. MineWorld \citep{guo2025mineworld} instead accelerates autoregressive discrete-token generation through parallel decoding, reporting up to 7 FPS in Minecraft. WHAM-RT combines one-shot frame prediction with lightweight masked refinement to achieve up to 17 FPS.}


\section{Preliminaries}

\noindent\textbf{Bleeding Edge.}\footnote{Official game website: \href{https://www.bleedingedge.com/en}{https://www.bleedingedge.com/en}} The approaches presented in this work are trained and evaluated on the Bleeding Edge environment. It is an action multiplayer game featuring 13 different playable characters and several different maps, providing a challenging testbed for world modelling due to its dynamic gameplay, multi-agent environments, various environmental hazards and interactive objects. 
Each of the characters has their own unique abilities, special attacks and superpowers, with 10 possible actions beyond navigation and camera control.  The data was collected in partnership with the Ninja Theory game studio, with the players' consent, and it amounts to roughly 500,000 anonymised gaming sessions. Image data was rendered in MP4 format at $60$ fps with a resolution of $300\times180$, alongside binary files containing the associated controller actions. A timecode extracted from the game was stored for each frame, to ensure actions and frames remained in sync at training time.


\noindent\textbf{WHAM} \citep{Kanervisto2025} is an autoregressive causal transformer model trained on this large dataset of human gameplay data.
Given a discretised series of interleaved tokens for video frames and corresponding controller actions, the model learns to predict the tokens of the next video frame and action in the sequence. The model uses a VQ-GAN \citep{esser2021taming} encoder/decoder trained from scratch on the same Bleeding Edge dataset and uses learned position embeddings.
Controller actions are encoded using a simple action tokeniser into two sets of tokens: a set of multi-hot encodings of button presses, and discretised $(x,y)$ coordinates of the controller joysticks that control the player and camera direction.

WHAM has demonstrated an ability to produce impressively consistent long sequences of gameplay, including character-specific abilities and interactions with non-playable characters. However, generating each frame is a slow process, making it unwieldy to use in practice in an interactive, playable setting, where the user will expect an instant reaction to their inputs.
In this work, we introduce WHAM-RT (WHAM Real-Time), which uses an alternative architecture in order to offer a similar experience to WHAM, but with much faster inference, thus enabling a truly responsive experience.
This improved generation speed is a crucial development if these models are to be used in practical applications such as prototyping in the future.

\section{Methods}

\subsection{WHAM Real-Time}
In order to accelerate the inference process of WHAM-like autoregressive models, we propose replacing the next-token prediction paradigm~\citep{Kanervisto2025} with an iterative masked prediction inference process, inspired by the \changed{MaskGIT} \citep{maskgit} formulation for image generation. In contrast to predicting one image token at a time, \changed{MaskGIT} aims to iteratively predict a subset of masked image tokens, starting from a fully masked image and iteratively reducing mask coverage until the full set of visual tokens is recovered. This strategy substantially accelerates inference, and was successfully leveraged in large foundation models for fast image \citep{muse_img_gen} and video generation \citep{phenaki}. Despite such speed gains, the foundation \changed{MaskGIT} models still require multiple iterations of prediction through a large transformer model, preventing them from reaching real-time performance.

\begin{figure}
    \centering
    \includegraphics[width=1\linewidth]{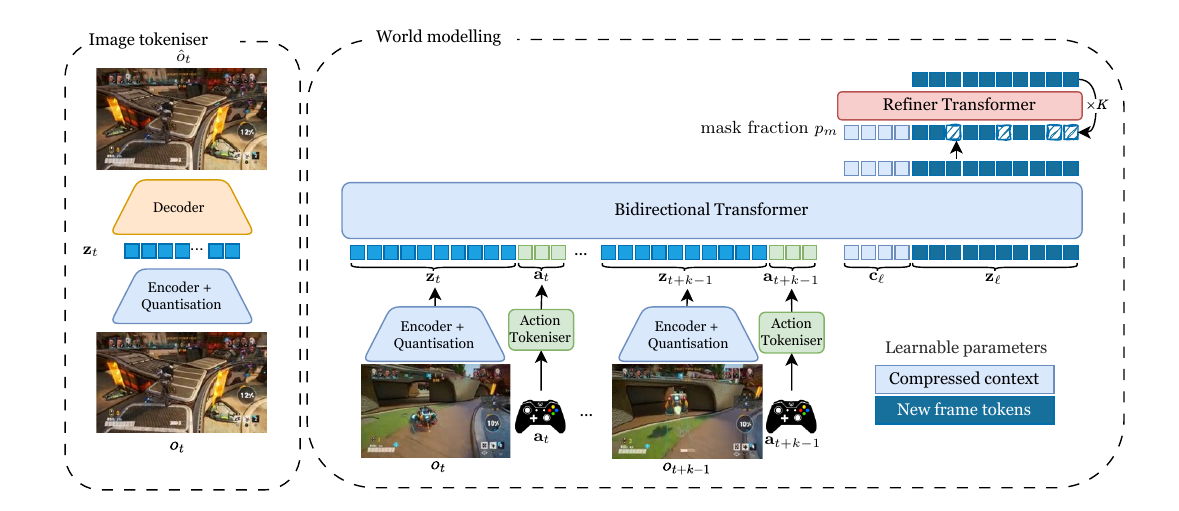}
    \captionsetup{font=changed}
    \caption{\changed{WHAM-RT architecture. A large backbone transformer predicts all next-frame image tokens in one shot and compresses the context. A lightweight discrete diffusion transformer refines the prediction using the compressed context.}}
    \label{fig:whamrt}
\end{figure}

\noindent\textbf{Model architecture.}
We design WHAM-RT to address these limitations, aiming to strike a balance between leveraging the predictive power of large autoregressive transformers and the benefits of iterative masked refinements, while at the same time maximising inference speed. Illustrated in Figure \ref{fig:whamrt}, WHAM-RT takes as input context a sequence $S$ of $N$ interleaved image and action tokens, tokenised following \cite{Kanervisto2025}. \changed{At frame $t$, a pre-trained VQ-GAN \citep{esser2021taming} encodes observation $o_t$ as discrete image tokens $\mathbf{z}_t$ and decodes them to reconstruction $\hat{o}_t$. Context $S$ spans $k$ frames with corresponding actions $\mathbf{a}_t$.} The model architecture comprises a large backbone transformer \changed{$\mathcal{B}$} that models temporal dependencies and generates a first estimate of the tokens of the next image in the sequence. Our backbone transformer has bidirectional attention, and predicts all new image tokens in one-shot \emph{in parallel}. To achieve real-time inference, iterative masked refinements are carried out using a lightweight refinement transformer \changed{$\mathcal{R}$}, which iteratively updates the backbone prediction.

\noindent\textbf{Video Context Compression.} In order to enable a one-shot image draft generation with 
the backbone transformer, we introduce a set of learnable input tokens \changed{$T_\ell = \{\mathbf{c}_\ell,\mathbf{z}_\ell\}$}, where \changed{$\mathbf{z}_\ell$ denotes learnable image-query tokens}, such that \changed{$\mathbf{z}_b = \mathcal{B}(\mathbf{z}_\ell | S, \mathbf{c}_\ell)$} are predicted image tokens of the next frame in sequence $S$. Tokens \changed{$\mathbf{c}_\ell$} play a crucial role towards achieving real-time performance of learning a compressed representation of context $S$. Similar to the concept of Registers (\texttt{REG}) in vision transformers \citep{darcet2024vision}, 
 \changed{$\mathbf{c}_\ell$} are a new set of learnable tokens introduced to summarise and compress the provided context.  
 This enables us to provide long context information to the refinement head while enabling fast refinement iterations. 
Concretely, we compute the compressed context as \changed{$\mathbf{c}_b = \mathcal{B}(\mathbf{c}_\ell | S, \mathbf{z}_\ell)$}. The masked iterative refinement is then carried out as \changed{$\mathbf{z}_r = \mathcal{R}(\mathbf{z}_b | \mathcal{M}_p, \mathbf{c}_b)$}, where \changed{$\mathcal{M}_p$ masks a fraction $p$ of image tokens}.

\noindent\textbf{Training.} We train the model by computing the cross-entropy loss between predictions from both backbone and the refinement transformer and the ground truth image tokens \changed{$\mathbf{z}$}. For the refinement transformer, we adopt a teacher forcing approach where ground truth tokens are provided as input. We randomly mask a \changed{fraction} $p$ of tokens such that $0 <  p \leq p_{max}$ and compute the loss of masked tokens only. As \cite{Kanervisto2025} demonstrated the benefits to predicting images and actions together, and to ensure training procedures are as similar as possible, we further introduce an action prediction auxiliary loss using a separate lightweight transformer head \changed{$\hat{\mathbf{a}} = \mathcal{A}(\mathbf{z}, \mathbf{c}_b)$}. Defining \changed{$\mathbf{a}$} as the ground truth action, our overall loss function is: 
\begin{equation}
    \changed{\mathcal{L} = \mathcal{L}_{CE}(\mathbf{z},\mathbf{z}_b) + \mathcal{L}_{CE}(\mathbf{z},\mathcal{R}(\mathbf{z}\mid \mathcal{M}_p,\mathbf{c}_b)) + \mathcal{L}_{CE}(\mathbf{a},\hat{\mathbf{a}})}
\end{equation}

\changed{Here $\mathcal{L}_{CE}$ denotes cross-entropy; the refinement term is evaluated only at masked positions. The refiner receives unmasked ground-truth tokens during training and model predictions at inference. We use discrete diffusion to refer to MaskGIT-style masking and iterative token recovery.}

\noindent\textbf{Inference.} \changed{Generation is autoregressive across frames. The backbone runs once on context $S$ to obtain compressed context $\mathbf{c}_b$ and draft $\mathbf{z}_b$, sampled at backbone temperature $\mathcal{T}_b$. We then perform $K$ refinement iterations. At each step $m=0,\ldots,K-1$, we mask a fraction $p_m=p_{max}(1-m/K)$ of the current image tokens, selecting positions uniformly at random, with $p_{max}=0.55$. The refiner resamples only masked tokens at refinement temperature $\mathcal{T}_r$, conditioned on the unmasked tokens and $\mathbf{c}_b$. With $K=0$, we use the backbone draft directly. The generated frame and its corresponding controller action update the rolling context for the next frame.}


\subsection{Training data efficient models}

Training game-specific world models usually focuses on large scale training with models like  DreamerV4~\citep{dreamerv4} and MineWorld~\citep{guo2025mineworld} using the VPT dataset~\citep{vpt}, which consists of ~8 years of gameplay. Similarly, our full WHAM-RT model was trained on the Skygarden bleeding edge dataset comprising 1 year of gameplay. Collecting sufficient gameplay data, and training models on such vast amounts of data can be prohibitively expensive and time consuming. This can restrict creatives in building and quickly testing custom environments. This section explores how, in a specific environment, we can \changed{train the dynamics model using} approximately a single day of recorded data by intentionally curating a high quality small dataset.

\begin{figure}
    \centering
    \includegraphics[angle=90,width=0.48\linewidth]{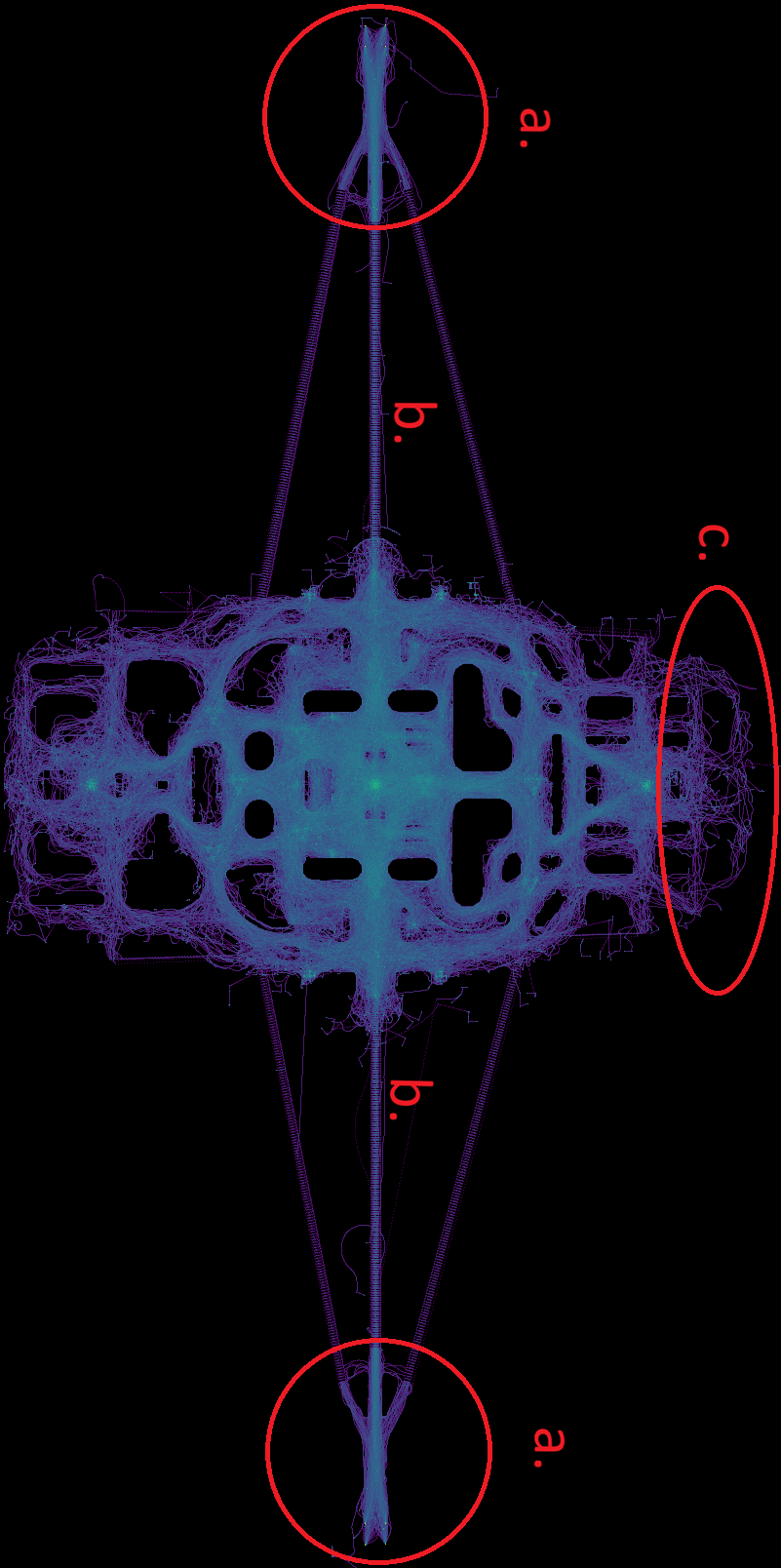}
    \includegraphics[angle=90,width=0.48\linewidth]{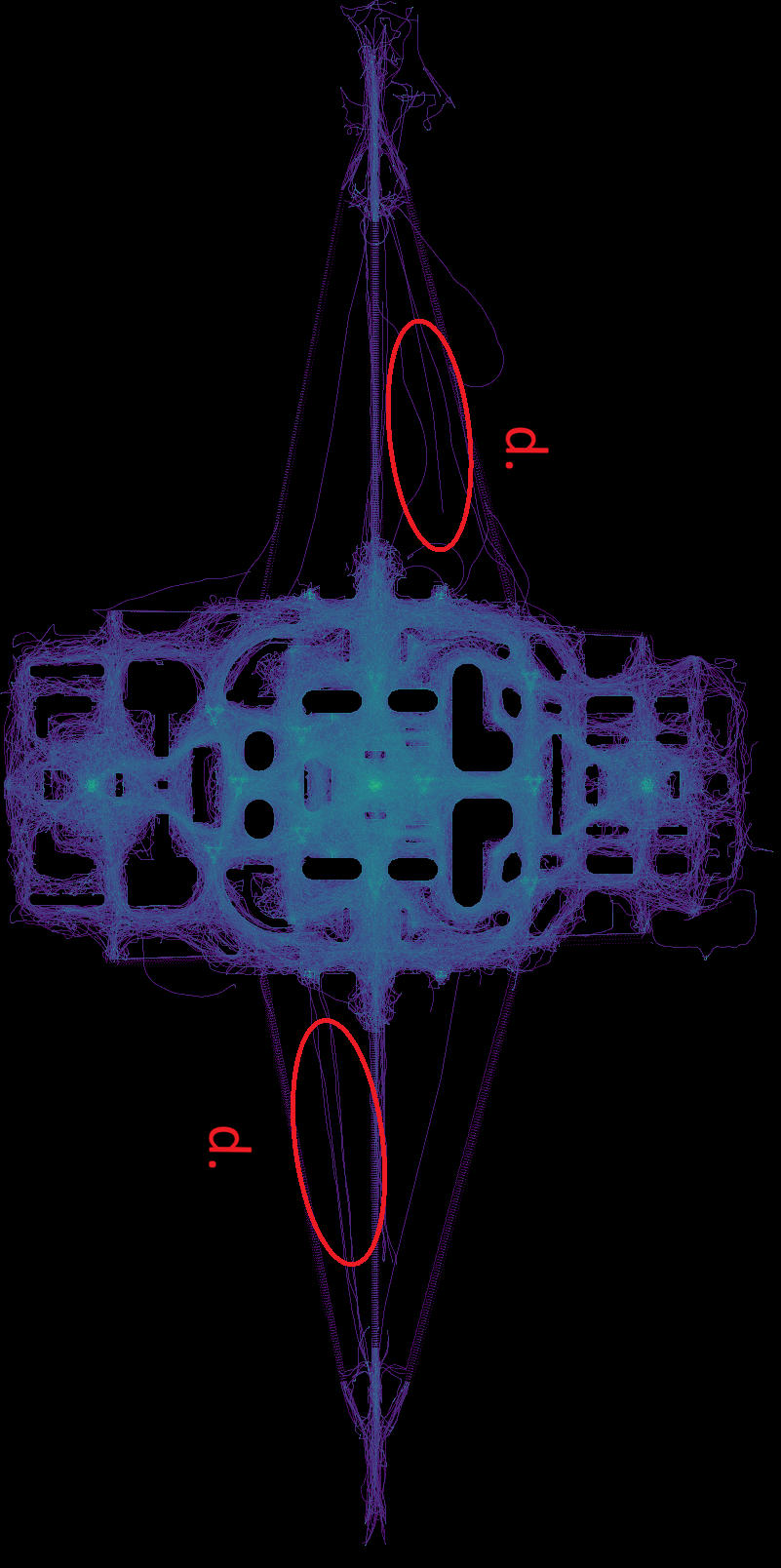}
    \caption{Heatmaps of agent locations on the sky garden map. Left: random sampling of 1000 sequences of 100 seconds each. Right: our curated dataset. The left image shows a bias towards the spawn points (a) and the corridors leading from them (b), and patchy coverage in the less frequently visited parts of the map, particularly the top and bottom extremities (c). Finally, the curated data captures rare occurrences, such as the extra paths between the spawn points and the main area (d).}
    \label{fig:data_curation}
\end{figure}

\noindent\textbf{Data Curation.} In addition to restricting the data to just one character, our approach to data curation focused on improving map coverage while countering biases in the natural human gameplay data. The full data contained a heavy bias towards certain parts of the map, particularly the player spawn points and the narrow corridors leading to the main area. Naively sampling the data leads to a dataset which contains redundant data in the frequently visited areas, while missing the lesser-visited parts of the map entirely. To counter this, we adopted an approach which considered every location visited by a trajectory in the full data, quantised these locations into a three-dimensional grid, 
then sampled each populated cell once, to give us a trajectory and time-step at which the player reached that position. In this way we chose $1000$ points uniformly spread across the playable area. The final curated dataset was formed by treating these as starting points and retrieving the subsequent hundred seconds of gameplay from their corresponding trajectories, to total $1000$ sequences $\times$ $100$s $\times$ $10$ FPS = one million steps. This does not result in a completely balanced dataset (where each spot on the map is visited equally frequently), but does limit the biases present in the full dataset, and reduces sparseness in the data. It also captures rare events, such as hard to reach locations of the map, which might only appear once in the full data and would be missed by naive sampling. This can be seen in Figure \ref{fig:data_curation}.

\section{Experiments}

\noindent\textbf{Implementation details.}
We reuse WHAM\changed{\textquoteright}s VQGAN tokeniser and GPT\changed{\mbox{-}}2\changed{--}style architecture for backbone and refinement transformers to ensure a fair comparison.
WHAM-RT's backbone and refinement transformers use a GPT2 architecture \citep{gpt2} with $24$ attention heads and embeddings of dimension $1536$. The former comprises $16$ layers for a total of $450$M parameters, while the latter has $8$ layers and $225$M parameters. The auxiliary head for action prediction uses a GPT2 causal transformer with $4$ layers. The compressed context tokens comprise $512$ tokens. We set $p_{max}=0.55$ for refinement iterations. The model has a context of 9 interleaved images ($540$ tokens per image) and actions ($16$ tokens per action). Models are trained on the Bleeding Edge dataset using the preprocessing and evaluation protocol from \cite{Kanervisto2025}. WHAM-RT was trained on the \texttt{Skygarden} game map, a set of $66$,$709$ individual player trajectories with an average length of around nine minutes each. After applying an $80$:$10$:$10$ split for training / validation / testing, and downsampling to $10$Hz, this gave us approximately $310$M frames (about one year of gameplay). Our standard WHAM-RT model comprises $741$ million parameters. For low data models, we additionally train a small model with 343 million parameters. We train the full dataset model for $200k$ steps with mixed precision using FSDP \citep{FSDP}, batch size $600$ and a learning rate of $4e^{-4}$ with $2000$ warm up steps and a square-root cooldown starting at $100k$ steps. The low data models are trained for $70k$ steps and batch size $240$, we use a learning rate of $3e^{-4}$ with $1000$ warm up steps and cooldown starting at step $50k$. The \changed{smaller model} comprises $16$ attention heads and embeddings of dimension $1024$, with other parameters identical. The full model is trained on 200 A100 GPUs for 3 days, low data models were trained on 128 A100 GPUs for 14h (\changed{standard}) and 9h (\changed{smaller}).


\begin{figure}[t]
    \centering
    \includegraphics[width=0.98\linewidth]{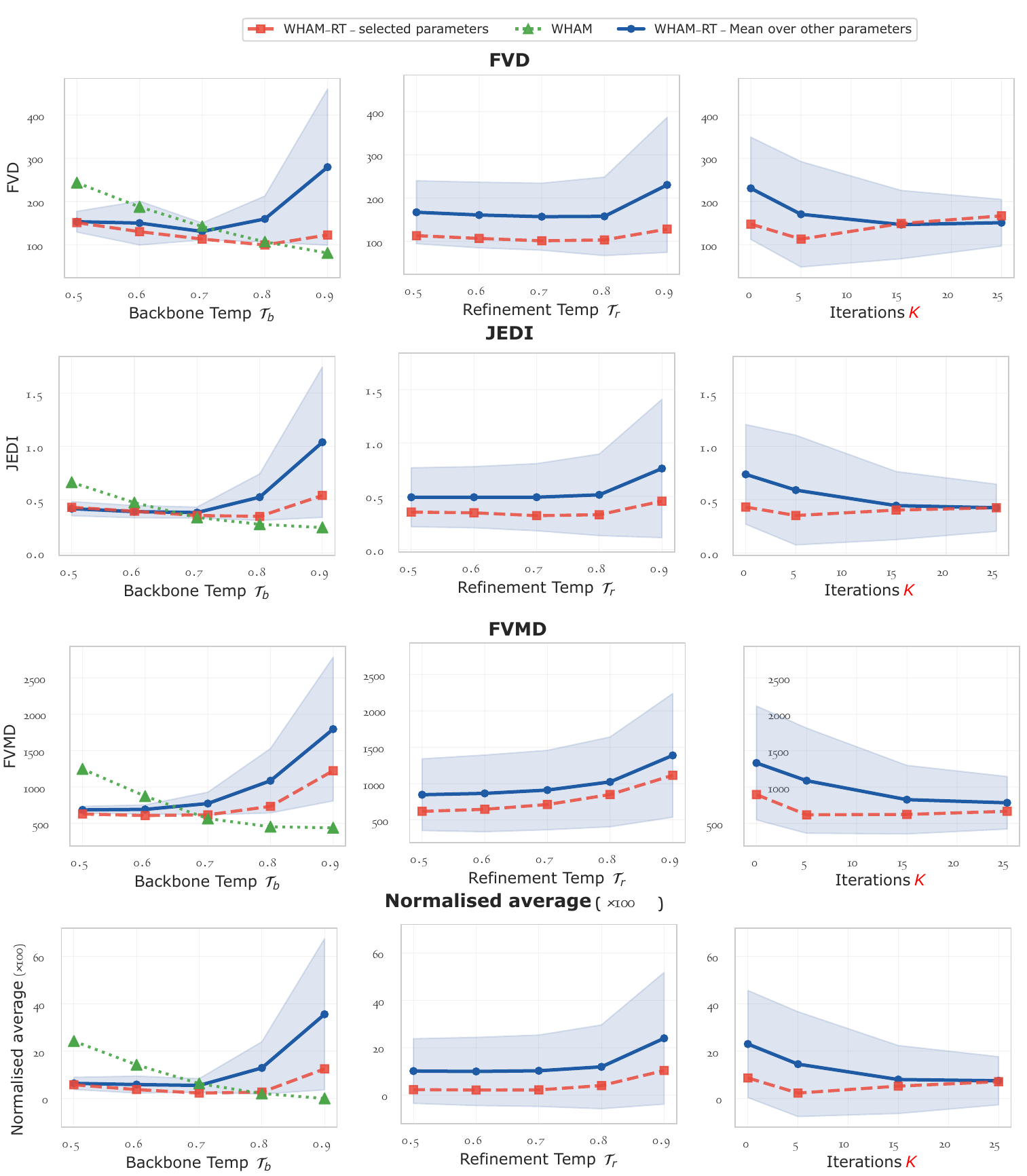}
    \caption{Impact of temperatures and iterations on FVD, FVMD, \changed{JEDI} and normalised average values. For a given parameter and metric, we report mean and standard deviation values with respect to other parameters. \changed{} We additionally show values for our default parameters (red line), and WHAM (green line). For all metrics, lower is better.}
    \label{fig:metrics}
\end{figure}


\noindent\textbf{WHAM-RT quantitative evaluation.} To evaluate WHAM-RT's performance, we focus our evaluation on 1) the quality of generated gameplay and 2) generation speed by measuring Frames Per Second (FPS). For quality, we compute three different metrics, measuring different aspects of generation performance. Firstly, similar to \cite{Kanervisto2025}, we compute the Fr\'{e}chet Video Distance (FVD) \citep{fvd}, together with a more robust variant, \changed{JEDI} \citep{jedi}, as a measure of statistical similarity between distributions of generated videos and ground truth. We additionally measure motion accuracy by computing the Fr\'{e}chet Video Motion Distance (FVMD) \citep{fvmd}, which compares motion vectors between ground truth and generated results. The latter was reported to be closer to human judgement than FVD-type metrics in \cite{gaia}.

In Figure \ref{fig:metrics}, we can see all metrics, compared to WHAM, for different values of backbone temperature $\mathcal{T}_b$. We can see that while WHAM consistently improves when $\mathcal{T}_b$ increases, WHAM-RT favours lower temperatures, with more instability at high temperatures. Results show that WHAM-RT's performance, in terms of consistency (FVD, \changed{JEDI}) shows a small performance loss compared to WHAM's best configuration. More differences are observed for FVMD, suggesting reduced motion consistency. This could be attributed to the compressed context used to refine visuals, potentially introducing jitters. \changed{}

\begin{figure*}[t]
    \centering

    \begin{minipage}[t]{0.48\linewidth}
        \centering
        \includegraphics[height=4.5cm, keepaspectratio]{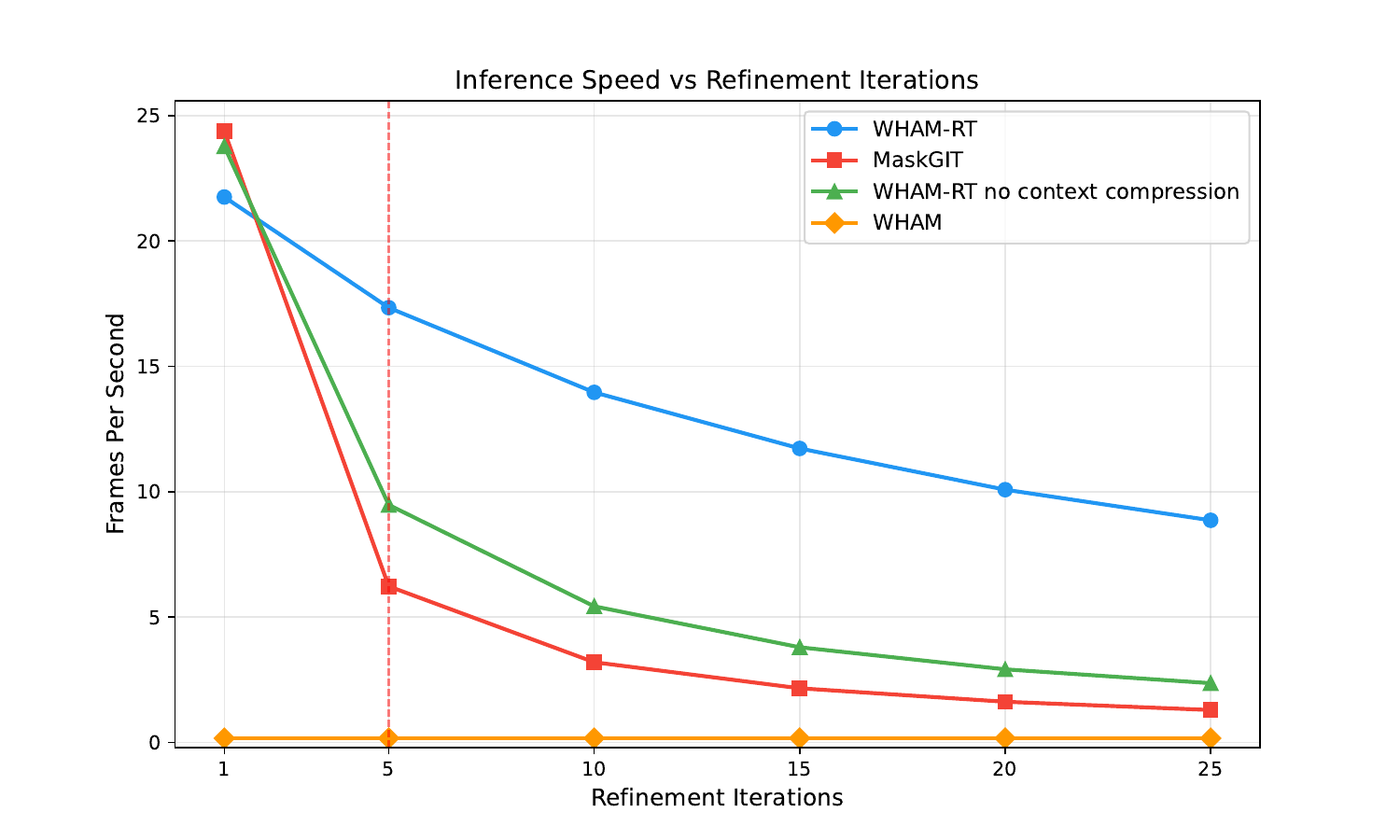}
        \caption{Comparison of model FPS during inference with respect to the number of refinement iterations calculated using a batch size of 1 on a single H100 accelerator.}
        \label{fig:fps}
    \end{minipage}
    \hfill
    \begin{minipage}[t]{0.48\linewidth}
        \centering
        \includegraphics[height=4cm, keepaspectratio]{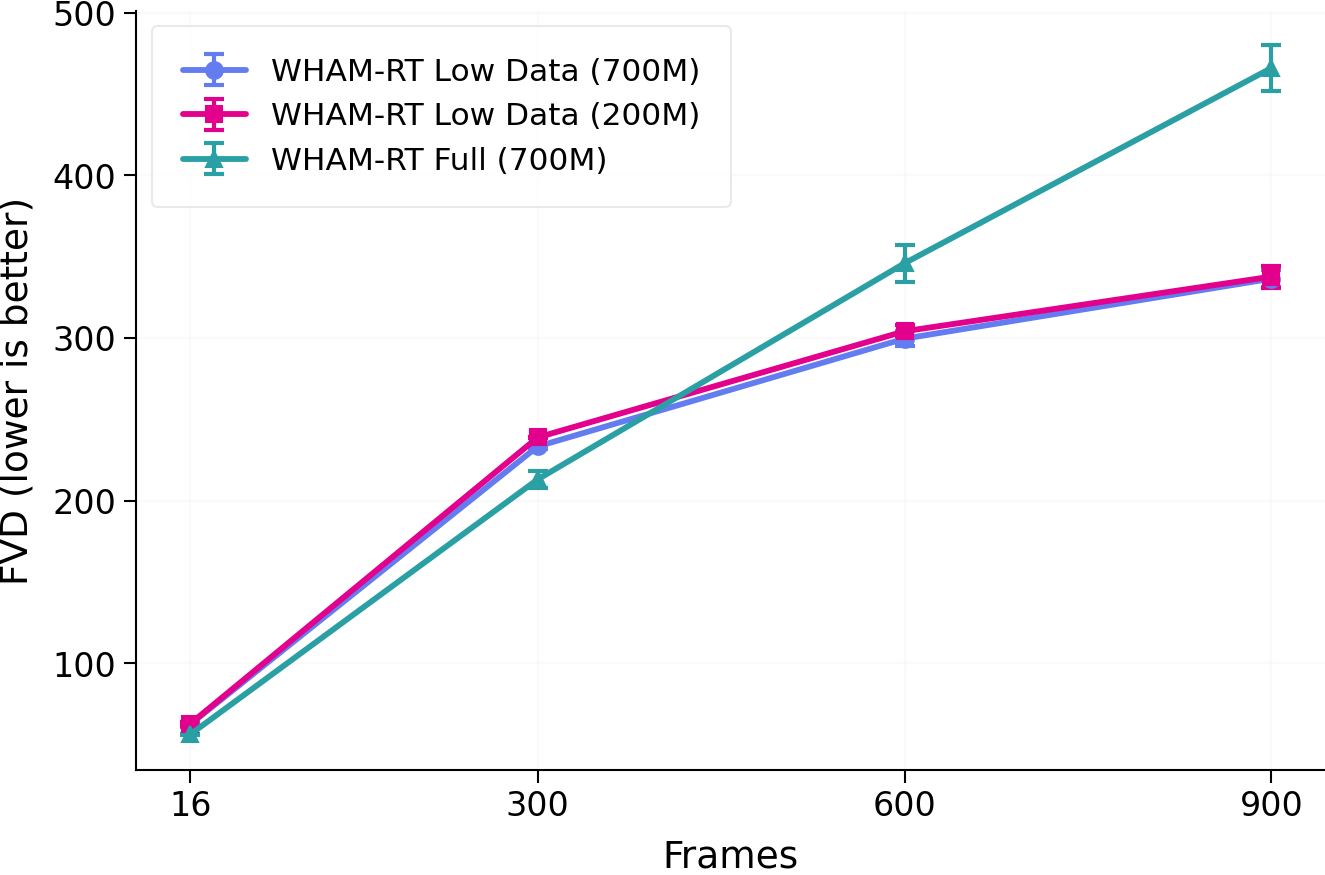}
        \caption{Low data models evaluation. We measure FVD over time for our full dataset model (WHAM-RT Full) and small data models at two model sizes (WHAM-RT Low Data).}
        \label{fig:fvd-1M}
    \end{minipage}
\end{figure*}

To measure generation speed, we compute FPS for different numbers of refinement iterations \changed{$K$}, as these directly impact speed. We compare to next token prediction (WHAM), \changed{MaskGIT} inference (i.e. carrying out refinement iterations within the backbone transformer), and WHAM-RT without context compression. We report results in Figure \ref{fig:fps}. We can see that WHAM-RT achieves substantial generation speedup reaching up to $17$ FPS with $5$ refinement iterations. This is an acceleration of up to 7895\% compared to WHAM, allowing users to interact with the model in real time. Compared to a \changed{MaskGIT} model, the acceleration is also significant. This  highlights that discrete diffusion is not sufficient for real-time inference, supporting the need for our two-stage formulation. Lastly, we note the impact of context compression on generation speed, allowing to substantially accelerate refinement iterations. 

\begin{figure}
    \centering
    \includegraphics[width=0.98\linewidth]{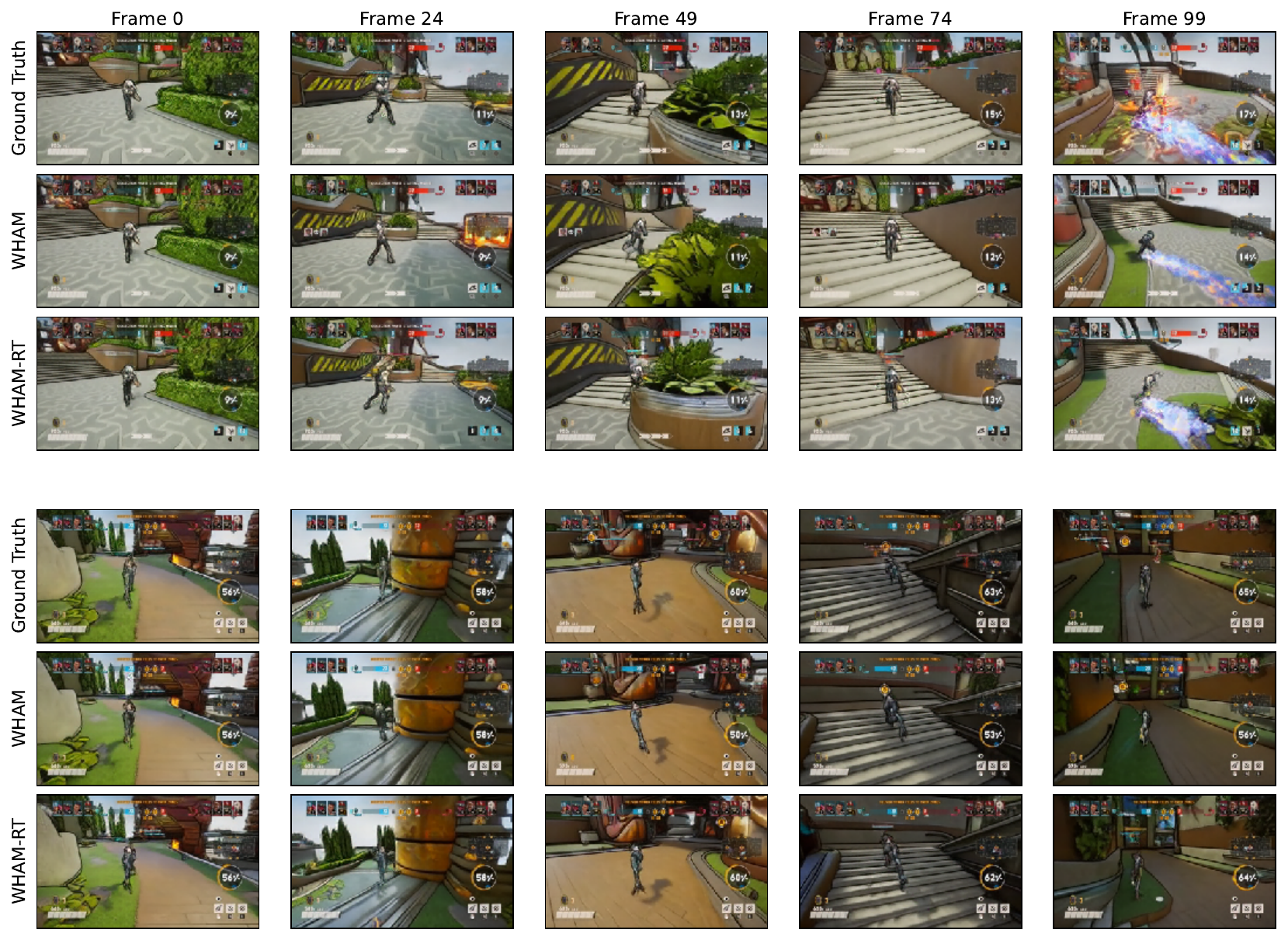}
    \caption{Visual comparisons of gameplay generated by WHAM-RT and WHAM, compared to the ground truth video. World models receive the first 10 ground truth frames as input, as well as the actions entered in the ground truth video.}
    \label{fig:qual_eval}
\end{figure}

\noindent\textbf{WHAM-RT qualitative evaluation.} We additionally provide a visual comparison between WHAM and WHAM-RT models in Figure \ref{fig:qual_eval}. We compare generated gameplay to the corresponding ground truth video which provides the context input (first 10 frames) and input actions. We can see very similar visual quality and environment consistency between models. We note that we expect slight differences between generated examples due to the stochastic nature of the environment.  

\noindent\textbf{Ablations.} 
In Table \ref{tab:git_token_ablation}, we provide inference-time ablations of key components of our WHAM-RT architecture, and report their impact on both speed (FPS) and world modelling quality (FVD). Similarly to prior quantitative results, we report FVD computed over our 1024 test examples. We compare our full WHAM-RT model to Backbone only (no refinement transformer, i.e. 0 iterations), WHAM-RT without compressed context (i.e. the refinement transformer has no context information) and WHAM-RT with compressed context from a random sequence. The latter aims to verify that the compressed context encodes sequence specific information instead of global environment context. 
Results show that WHAM-RT achieves a good balance between inference speed and generation quality, providing an improvement over backbone only prediction, while maintaining real-time inference performance. \changed{Removing the compressed context yields only a small speed increase, indicating that compressed conditioning adds little inference overhead.} Lastly, using a random context yields substantial performance degradation, demonstrating that the compressed context integrates information about the sequence.

In addition to the backbone temperature, we investigate the influence of two key parameters: the refinement temperature $\mathcal{T}_r$ and iterations \changed{$K$}. We generate gameplay sequences for all $1024$ test examples using $100$ different configurations, with \changed{$\mathcal{T}_r,\mathcal{T}_b \in [0.5,0.9]$} and \changed{$K \in \{0,5,15,25\}$}. Results in Figure \ref{fig:metrics} 
show that the backbone temperature has the largest influence over performance, with better quality results at low temperatures and \changed{the highest temperatures} substantially altering generation quality. The refinement temperature shows a similar, but more stable behaviour. \changed{Lastly, larger numbers of iterations yield results that are more robust to temperature values; however, fewer refinement iterations (with $K>0$) achieve the best overall performance at low temperatures.} \changed{This is likely because repeated low-temperature refinement may favour static, high-probability scenes over dynamic gameplay.}

\begin{table}[t]
\caption{WHAM-RT ablations. We compare full WHAM-RT with variants removing or perturbing the compressed context, as well as the backbone-only model. Lower FVD and higher FPS is better.}
\label{tab:git_token_ablation}
\centering
\begin{tabular}{lll}
\toprule
\textbf{Model} & \textbf{FVD} $\downarrow$ & \textbf{FPS} $\uparrow$ \\
\midrule
Backbone only (no refinement) & 147.7 & 21.7 \\
WHAM-RT (no context) & 2576.1 & 18.2 \\
WHAM-RT (random compressed context) & 825.5 & 17.3 \\
WHAM-RT & 115.0 & 17.3 \\
\bottomrule
\end{tabular}
\end{table}

\noindent\textbf{Long horizon coherence.}
\changed{Here w}e test WHAM-RT's ability to generate frames across an extended time horizon. \changed{For a quantitative comparison, we generate $930$-frame rollouts at $10$ FPS using held-out recorded actions. All models are evaluated on examples of the curated model's character. WHAM serves as an empirical quality ceiling that the faster WHAM-RT variants aim to approach. Table \ref{tab:additional_fvd} compares full-data WHAM-RT with two variants trained on curated or random $28$-hour subsets. On longer rollouts, both variants have lower FVD than full-data WHAM-RT, narrowing the gap to WHAM. This shows that smaller training sets can support stronger long-horizon coherence.}

\changed{For a qualitative assessment, two human participants each recorded five minutes of interactive WHAM-RT gameplay using a controller or keyboard. Figure \ref{fig:time} shows frames generated by the full-data model during these sessions.} The full videos are provided in the supplementary material.

\begin{table}[!htbp]
\centering
\changed{\captionsetup{font=changed}\caption{Long-horizon FVD ($\downarrow$). Both low-data WHAM-RT variants outperform the full-data model on longer rollouts.}\label{tab:additional_fvd}}
\changed{\begin{tabular}{r r r r r}
\toprule
Frames & WHAM & WHAM-RT full & $28$h curated & $28$h random \\
\midrule
$16$ & $48.6$ & $57.4$ & $61.9$ & $65.9$ \\
$300$ & $126.9$ & $211.3$ & $217.1$ & $230.8$ \\
$600$ & $189.4$ & $357.1$ & $268.4$ & $275.9$ \\
$900$ & $243.7$ & $488.5$ & $298.5$ & $297.7$ \\
\bottomrule
\end{tabular}}
\end{table}

\changed{\begin{figure}[!htbp]
    \centering
    \includegraphics[width=\linewidth]{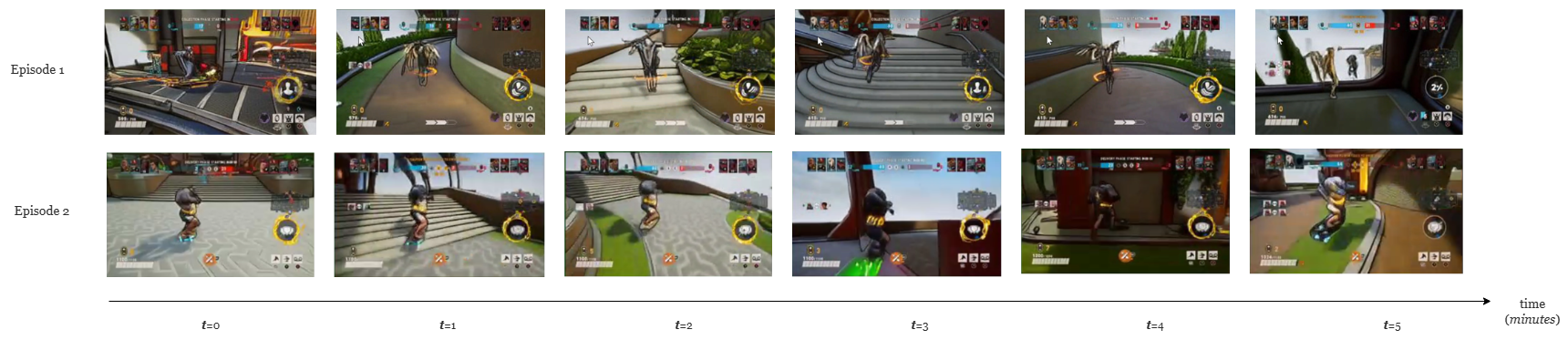}
    \captionsetup{font=changed}
    \caption{Generated frames taken each minute from 2 long gameplay sequences with WHAM-RT.}
    \label{fig:time}
\end{figure}}

\noindent\textbf{Low data models.}
\changed{In this section, we seek to evaluate the robustness of low data models, compared to full WHAM-RT. Figure \ref{fig:fvd-1M} reports comparative results between the full model, and standard and smaller variants of our low data models. At longer horizons, the low-data models have lower FVD than full WHAM-RT. We also report I3D feature-space density and coverage \href{https://proceedings.mlr.press/v119/naeem20a.html}{(Naeem et al., ICML 2020)}, with $k=5$ nearest neighbours. Density is the average number of reference neighbourhoods containing each generated sample, normalised by $k$; coverage is the fraction of reference samples whose neighbourhood contains a generated sample. In Table \ref{tab:curation_metrics}, the curated model has higher density and coverage beyond $16$ frames, while FVD is similar at $900$ frames. This motivates further study of small, high-quality datasets, as explored in language modelling \citep{gunasekar2023textbooksneed, zhou2023lima}.}

\begin{table}[!htbp]
\centering
\changed{\captionsetup{font=changed}\caption{The curated low-data model has higher feature-space density and coverage than the random-subset model on longer rollouts.}\label{tab:curation_metrics}}
\changed{\small\begin{tabular}{r c c c c}
\toprule
& \multicolumn{2}{c}{Density $\uparrow$} & \multicolumn{2}{c}{Coverage $\uparrow$} \\
Frames & Curated & Random & Curated & Random \\
\midrule
$16$ & $1.171\pm0.035$ & $1.212\pm0.034$ & $0.999\pm0.001$ & $0.999\pm0.001$ \\
$300$ & $0.659\pm0.007$ & $0.601\pm0.010$ & $0.749\pm0.005$ & $0.710\pm0.003$ \\
$600$ & $0.569\pm0.006$ & $0.494\pm0.003$ & $0.658\pm0.003$ & $0.612\pm0.002$ \\
$900$ & $0.525\pm0.005$ & $0.440\pm0.004$ & $0.612\pm0.001$ & $0.560\pm0.002$ \\
\bottomrule
\end{tabular}}
\end{table}


\changed{\noindent\textbf{Human evaluation.} Finally, we conduct a small human evaluation study to assess to what degree WHAM-RT approaches the quality of WHAM generation.
In a blinded study, six expert annotators each rated $30$ pairs of $11$-second clips from WHAM and full WHAM-RT, with presentation order randomised per trial. Of $180$ judgements, $80$ ($44.4\%$) preferred WHAM, $36$ ($20.0\%$) preferred WHAM-RT, and $64$ ($35.6\%$) found no discernible difference. Annotators highlighted WHAM's visual and character stability. Where WHAM-RT was preferred, annotators highlighted more defined combat interactions and more accurate modelling of mechanics such as jump-pad navigation and healing. Together with the inference-speed results, these findings show that WHAM-RT achieves substantial acceleration to real-time generation while retaining competitive perceptual quality: it was preferred or judged indistinguishable from WHAM in $55.6\%$ of judgements.}

\section{Conclusions}

In this work, we make advances in the efficiency of world models, with a focus on two key areas: training in a low-data scenario, and increasing inference speed while \changed{maintaining} a high level of quality and consistency.
Firstly, we introduce WHAM-RT, a novel world model architecture building on \changed{the} highly consistent world model WHAM \citep{Kanervisto2025}, enabling real-time inference through a novel two-stage modelling approach combining lightweight discrete diffusion with one-shot temporal modelling. 
Subsequently, we demonstrate that through a careful data sampling approach, we can successfully train this model on a fraction of the data ordinarily used to train similar world models.
These improvements in efficiency could open the door to applications of world models that may have previously been impractical.
We hope that these findings stimulate further work on high-quality world models with a focus on maximising efficiency.

\noindent \textbf{Limitations and future work.}
Our findings are limited to the Bleeding Edge environment to ensure \changed{a} fair and accurate comparison to our key baseline WHAM. Bleeding Edge constitutes a highly complex environment with \changed{in-depth} interactions (more than $10$ different types of actions) providing a highly challenging test bed for world models. Future work will investigate \changed{cross-environment} training and adaptation.
\changed{Achieving substantial acceleration did reduce world modelling quality compared to WHAM. Future work will explore improved model architectures, as well as newer image generation approaches \citep{VAR, MAR}, to further improve both generation quality and speed.}


\begin{ack}\changed{This work was carried out at Microsoft. We thank the Ninja Theory team for their collaboration on the Bleeding Edge gameplay dataset, Shanzheng Tan for his support with data and infrastructure, and the participants who contributed to our human evaluations.}\end{ack}

\bibliography{references}
\bibliographystyle{unsrtnat}

\newpage
\appendix

\section{Video results}
\label{supp:videos}
As supplementary material, we include representative generated video examples from \changed{the following models} used in our experiments:
\begin{itemize}
\item WHAM: 3 examples, length: 11 seconds each
\item WHAM-RT: 3 examples, length: 11 seconds each
\item WHAM-RT (long time horizon videos): 2 examples, length: 5 minutes each
\end{itemize}

\section{Additional Quantitative Results}
\label{supp:addquantresults}

\begin{figure}[!t]
    \centering
    \begin{subfigure}[t]{\linewidth}
    \centering\includegraphics[width=\linewidth]{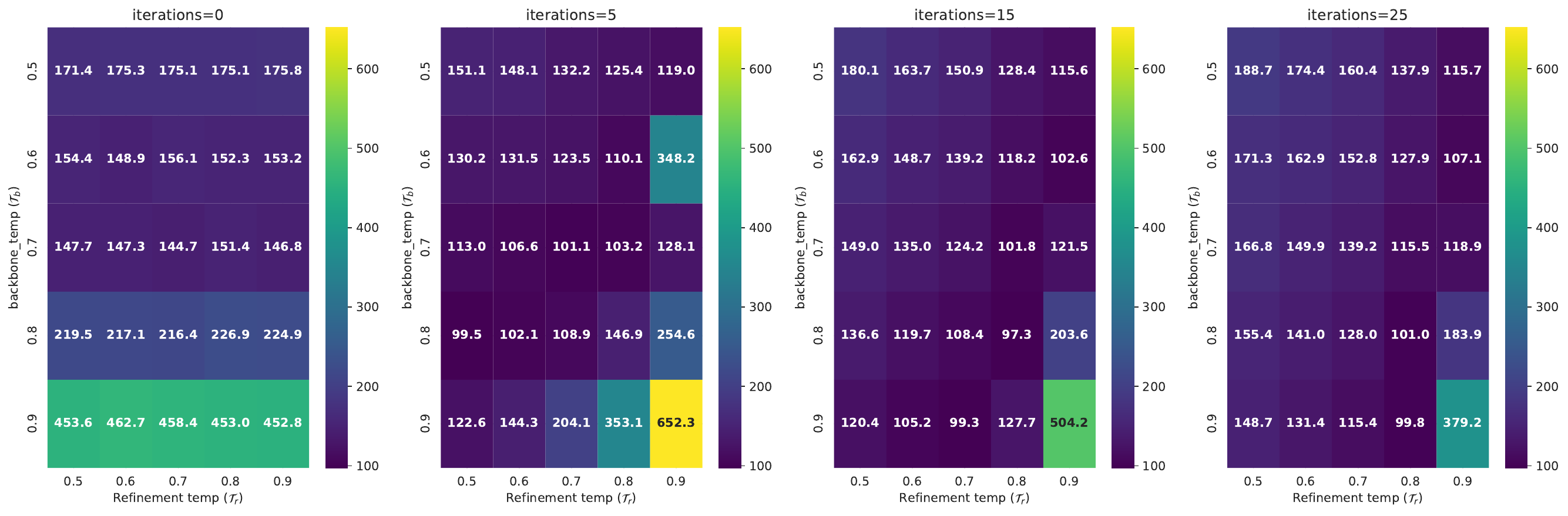}
    \caption{FVD}
    \end{subfigure}
      \begin{subfigure}[t]{\linewidth}
    \centering\includegraphics[width=\linewidth]{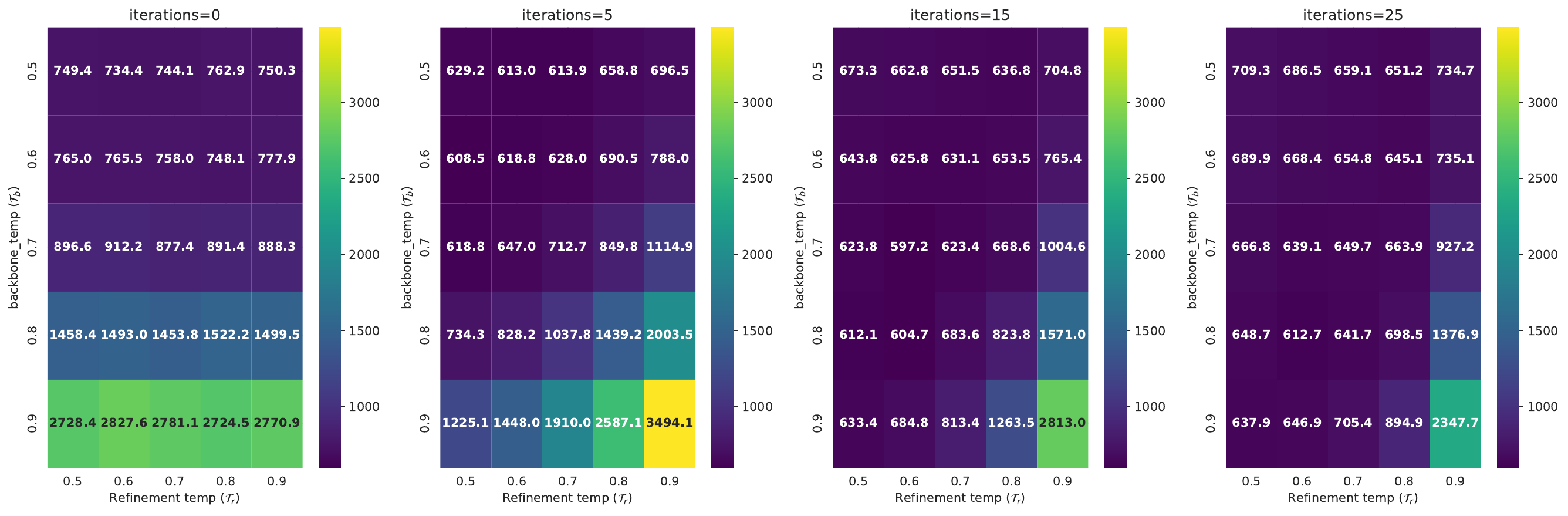}
    \caption{FVMD}
    \end{subfigure}
  \begin{subfigure}[t]{\linewidth}
    \centering\includegraphics[width=\linewidth]{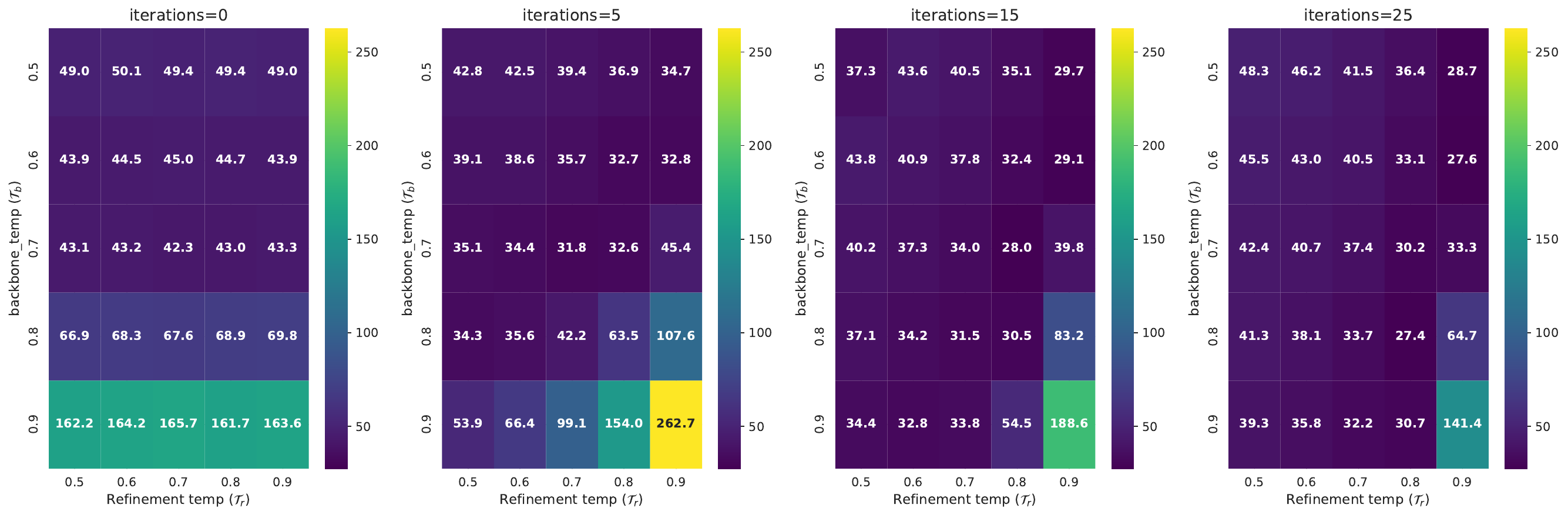}
    \caption{\changed{JEDI} ($\times 100$)}
    \end{subfigure}
\begin{subfigure}[t]{\linewidth}
\centering\includegraphics[width=\linewidth]{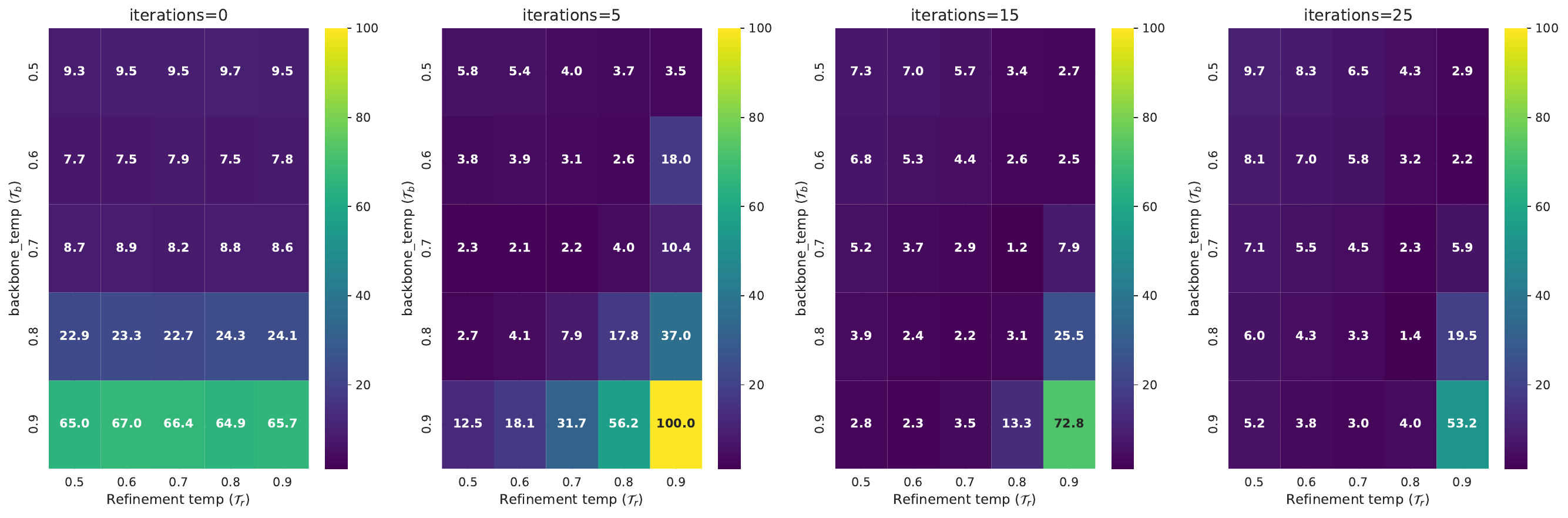}
\caption{Normalised average ($\times 100$)}
  \end{subfigure}
  \caption{Detailed quantitative results: heatmaps reporting computed metrics for all temperature and iteration settings.}
  \label{supp:heatmaps}
\end{figure}

\begin{figure}[t]
    \centering
    \includegraphics[width=\linewidth]{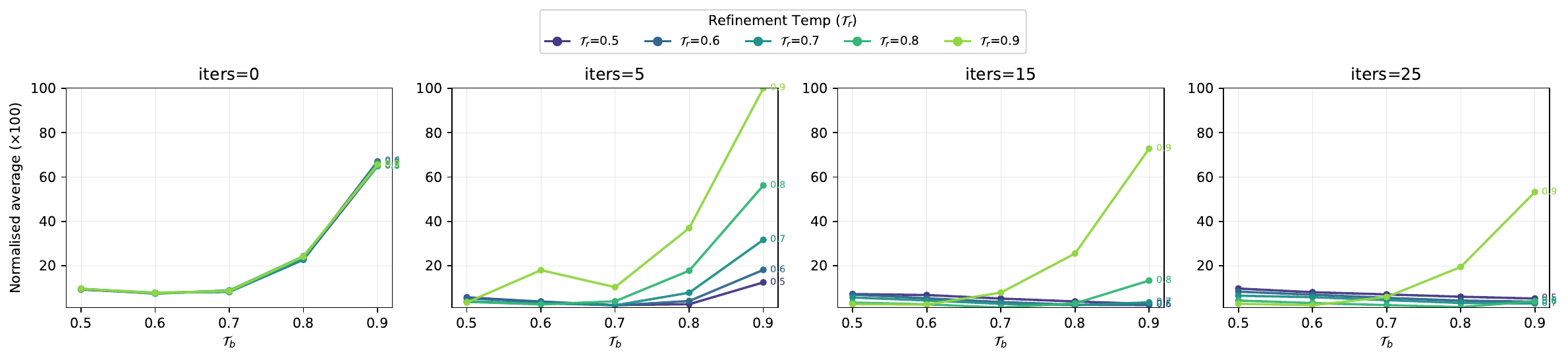}
    \caption{Impact of temperature parameters on the normalised average of computed metrics (FVD, FVMD, \changed{JEDI}) per number of iterations. Per metric results are available in Appendix \ref{supp:addquantresults}.}
    \label{fig:iters}
\end{figure}

\begin{figure}[t]
    \centering
    \begin{subfigure}[t]{\linewidth}
    \centering\includegraphics[width=\linewidth]{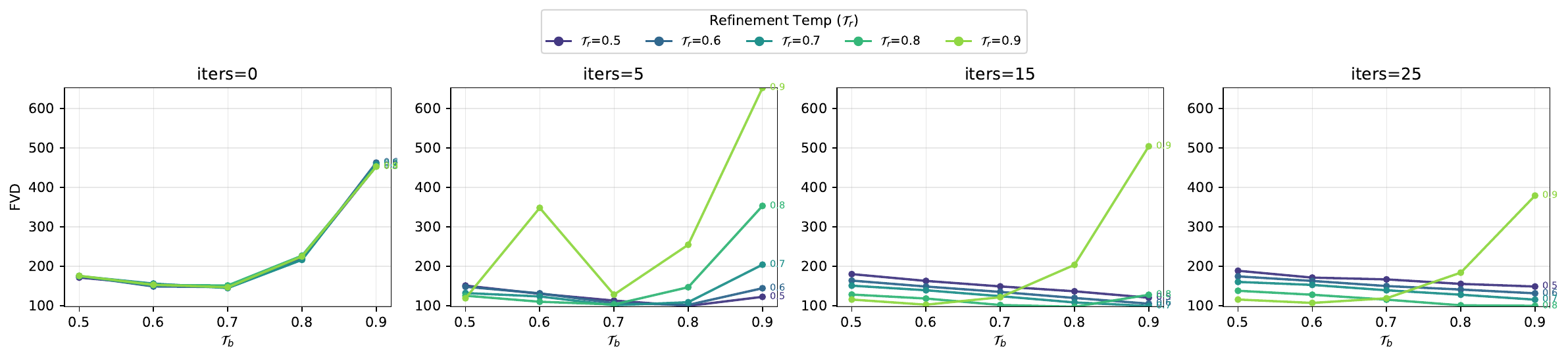}
    \caption{FVD}
    \end{subfigure}
      \begin{subfigure}[t]{\linewidth}
    \centering\includegraphics[width=\linewidth]{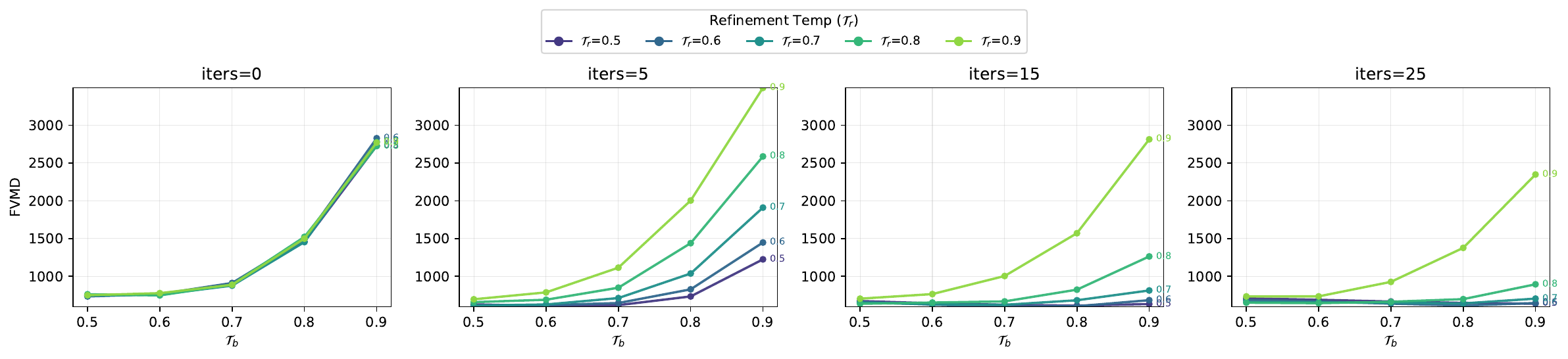}
    \caption{FVMD}
    \end{subfigure}
  \begin{subfigure}[t]{\linewidth}
    \centering\includegraphics[width=\linewidth]{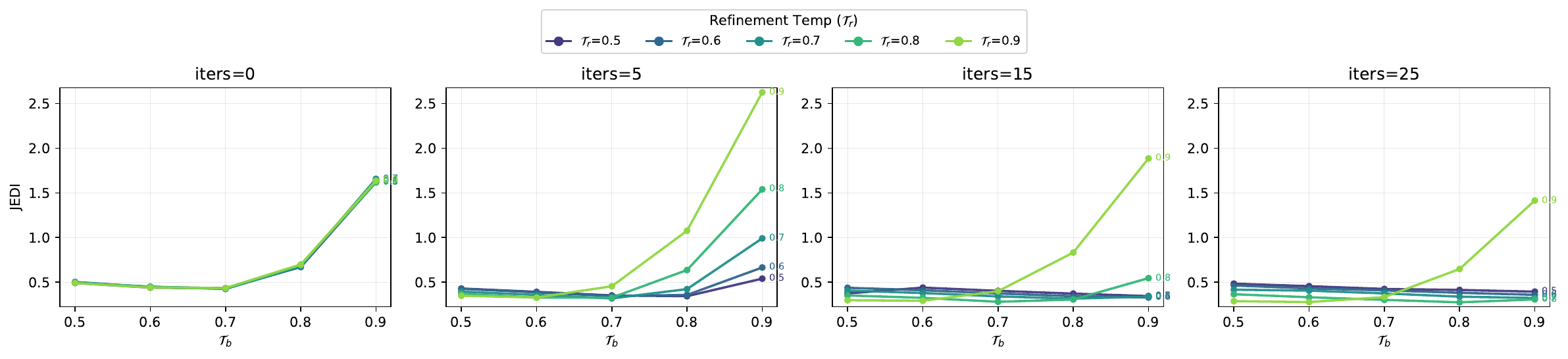}
    \caption{\changed{JEDI}}
    \end{subfigure}
  \caption{Per metric results. We show individual metric results and the impact of temperature and iteration values on overall performance.}
  \label{supp:iters}
\end{figure}

In this section, we provide detailed results for FVD, FVMD and \changed{JEDI} metrics, as well as their normalised average, computed by averaging normalised metrics. We provide iteration-stratified heatmaps in Figure \ref{supp:heatmaps}, showing metrics computed for all configurations. \changed{JEDI} and Normalised average values were multiplied by $100$ to improve readability. 
In addition to Figure \ref{fig:iters}, which reports results for the normalised average, we show the impact of refinement temperature and number of iterations over all individual metrics in Figure \ref{supp:iters}, with respect to the backbone temperature. 

We can see that all metrics become more stable at higher temperature with more iterations, with FVD and \changed{JEDI} preferring high refinement temperatures, while FVMD achieves consistently better results at low temperatures. This can be attributed to the fact that the metrics measure different properties of the video, with FVD/\changed{JEDI} measuring visual similarity, while FVMD measures motion accuracy. Visually, we have observed more lively environments (e.g. more NPCs) at high temperatures, at the expense of more flickering and instability, which supports these observations. 

\changed{\noindent\textbf{Long-horizon evaluation.} For Table \ref{tab:additional_fvd}, we evaluate $788$ held-out examples featuring the curated model's character, without restricting map location. Results for WHAM and full-data WHAM-RT use one evaluation seed; results for the two low-data variants are means over four evaluation seeds. Table \ref{tab:curation_metrics} reports means and standard deviations over four evaluation seeds.}

\section{Additional Qualitative Results}
\label{supp:addvisresults}
\begin{figure}
    \centering
    \includegraphics[width=\linewidth]{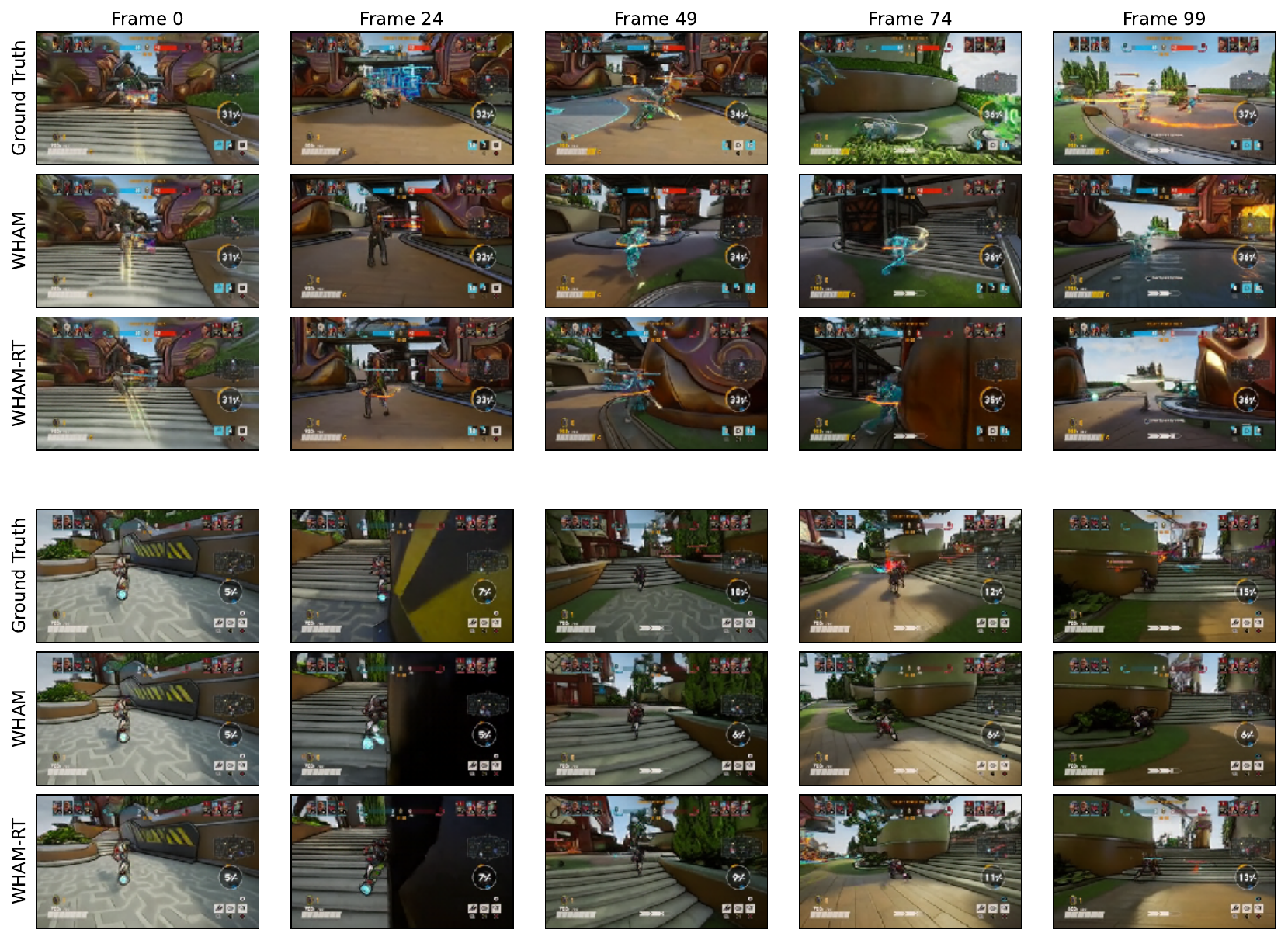}
    \caption{Additional visual results showing gameplay generated by WHAM-RT and WHAM, compared to the
ground truth video.}
    \label{fig:visresults2}
\end{figure}

\begin{figure}
    \centering
    \includegraphics[width=\linewidth]{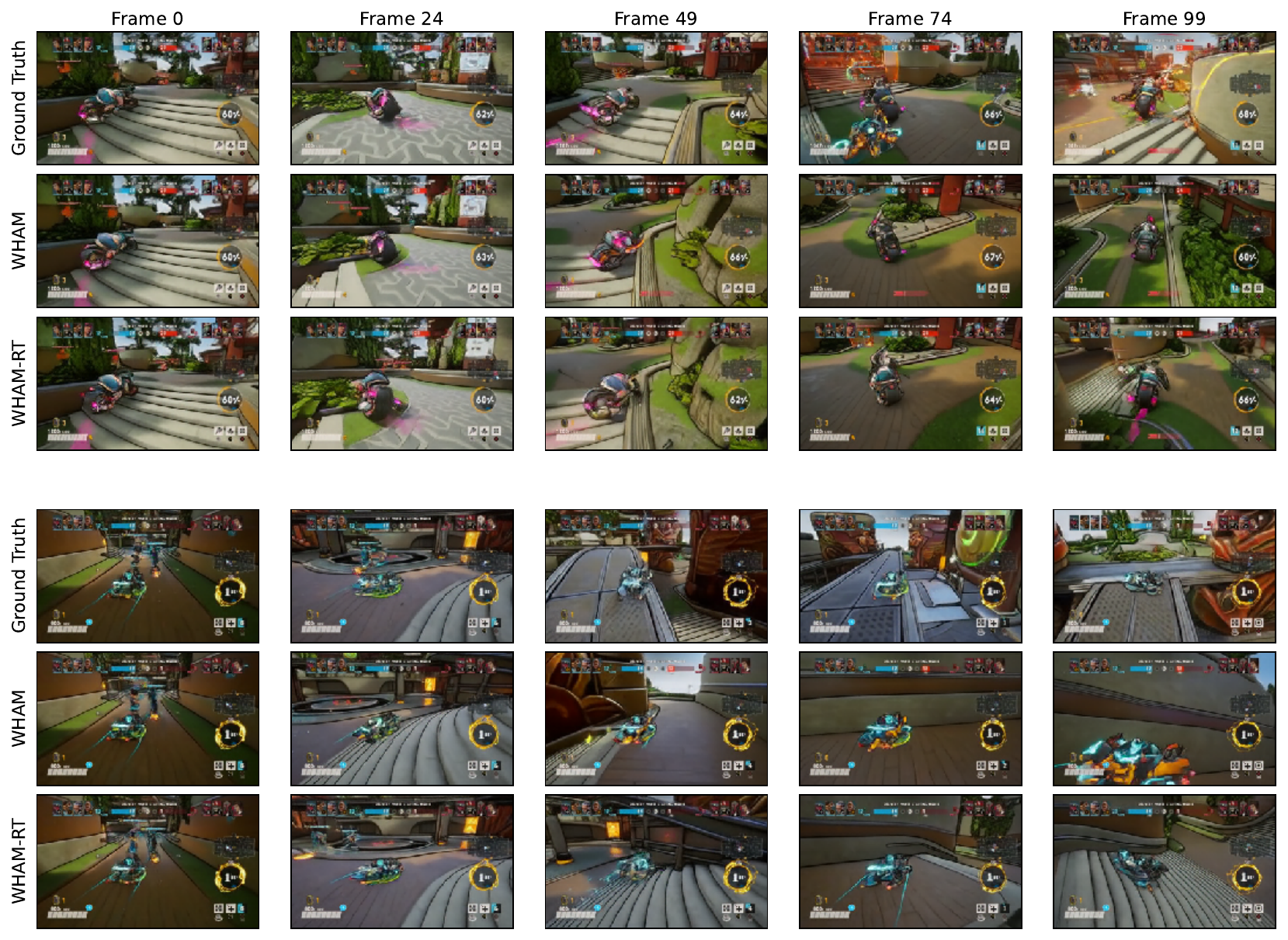}
    \caption{Additional visual results showing gameplay generated by WHAM-RT and WHAM, compared to the
ground truth video.}
    \label{fig:visresults3}
\end{figure}

 We provide additional visual results comparing WHAM and WHAM-RT to ground truth sequences in Figures \ref{fig:visresults2} and \ref{fig:visresults3}, showing very similar performance between WHAM and WHAM-RT models.

\end{document}